\documentclass[lettersize,journal]{IEEEtran}
\usepackage{amsmath,amsfonts}
\usepackage{algorithmic}
\usepackage{algorithm}
\usepackage{array}
\usepackage[caption=false,font=normalsize,labelfont=sf,textfont=sf]{subfig}
\usepackage{textcomp}
\usepackage{stfloats}
\usepackage{url}
\usepackage{verbatim}
\usepackage{graphicx}
\usepackage{cite}
\usepackage{multirow}
\usepackage[colorlinks,linkcolor=red,anchorcolor=blue,citecolor=blue]{hyperref}
\begin{document}

\title{Open-Vocabulary X-ray Prohibited Item Detection via Fine-tuning CLIP}

\author{Shuyang Lin, Tong Jia, Hao Wang, Bowen Ma, Mingyuan Li, and Dongyue Chen

\thanks{This work is supported by the National Natural Science Foundation of China (NSFC) under Grant 62173083, Grant U22A2063, and Grant 62206043; the National Key Research and Development Project of China under Grant 2022YFF0902401; Guangdong Basic and Applied Basic Research Foundation (2021B1515120064); the Major Program of National Natural Science Foundation of China (71790614), and the 111 Project (B16009). \textit{(Corresponding author: Tong Jia.)}}
\thanks{Shuyang Lin, Hao Wang, Bowen Ma, and Mingyuan Li are with the College of Information Science and Engineering, Northeastern University, Shenyang, 110819, Liaoning, China (e-mail: 2210329@stu.neu.edu.cn; ddsywh@yeah.net; 2010285@stu.neu.edu.cn; 542027743@qq.com).}
\thanks{Tong Jia is with the College of Information Science and Engineering, Northeastern University, Shenyang, 110819, Liaoning, China, and also with the Key Laboratory of Data Analytics and Optimization for Smart Industry, Ministry of Education, Northeastern University, Shenyang, 110819, Liaoning, China (e-mail: jiatong@ise.neu.edu.cn).}
\thanks{Dongyue Chen is with the College of Information Science and Engineering, Northeastern University, Shenyang, 110819, Liaoning, China, and also with the Foshan Graduate School of Innovation, Northeastern University, Foshan, 528311, Guangdong, China (e-mail: chendongyue@ise.neu.edu.cn).}}

\markboth{Journal of \LaTeX\ Class Files,~Vol.~14, No.~8, August~2021}%
{Shell \MakeLowercase{\textit{et al.}}: A Sample Article Using IEEEtran.cls for IEEE Journals}


\maketitle

\begin{abstract}
X-ray prohibited item detection is an essential component of security check and categories of prohibited item are continuously increasing in accordance with the latest laws. Previous works all focus on close-set scenarios, which can only recognize known categories used for training and often require time-consuming as well as labor-intensive annotations when learning novel categories, resulting in limited real-world applications. Although the success of vision-language models (e.g. CLIP) provides a new perspectives for open-set X-ray prohibited item detection, directly applying CLIP to X-ray domain leads to a sharp performance drop due to domain shift between X-ray data and general data used for pre-training CLIP. To address aforementioned challenges, in this paper, we introduce distillation-based open-vocabulary object detection (OVOD) task into X-ray security inspection domain by extending CLIP to learn visual representations in our specific X-ray domain, aiming to detect novel prohibited item categories beyond base categories on which the detector is trained. Specifically, we propose X-ray feature adapter and apply it to CLIP within OVOD framework to develop OVXD model. X-ray feature adapter containing three adapter submodules of bottleneck architecture, which is simple but can efficiently integrate new knowledge of X-ray domain with original knowledge, further bridge domain gap and promote alignment between X-ray images and textual concepts. Extensive experiments conducted on PIXray and PIDray datasets demonstrate that proposed method performs favorably against other baseline OVOD methods in detecting novel categories in X-ray scenario. It outperforms previous best result by 15.2 AP$_{50}$ and 1.5 AP$_{50}$ on PIXray and PIDray with achieving 21.0 AP$_{50}$ and 27.8 AP$_{50}$ respectively. Furthermore, our model can be directly transferred to different X-ray datasets without fine-tuning, verifying its generalization ability.
\end{abstract}

\begin{IEEEkeywords}
X-ray security inspection, prohibited item detection, open-vocabulary object detection, vision-language model, adapter-based fine-tuning.
\end{IEEEkeywords}

\section{Introduction}
\label{sec:introduction}
\begin{figure}[t]
    \centering
    \includegraphics[height=6cm,width=8.8cm]{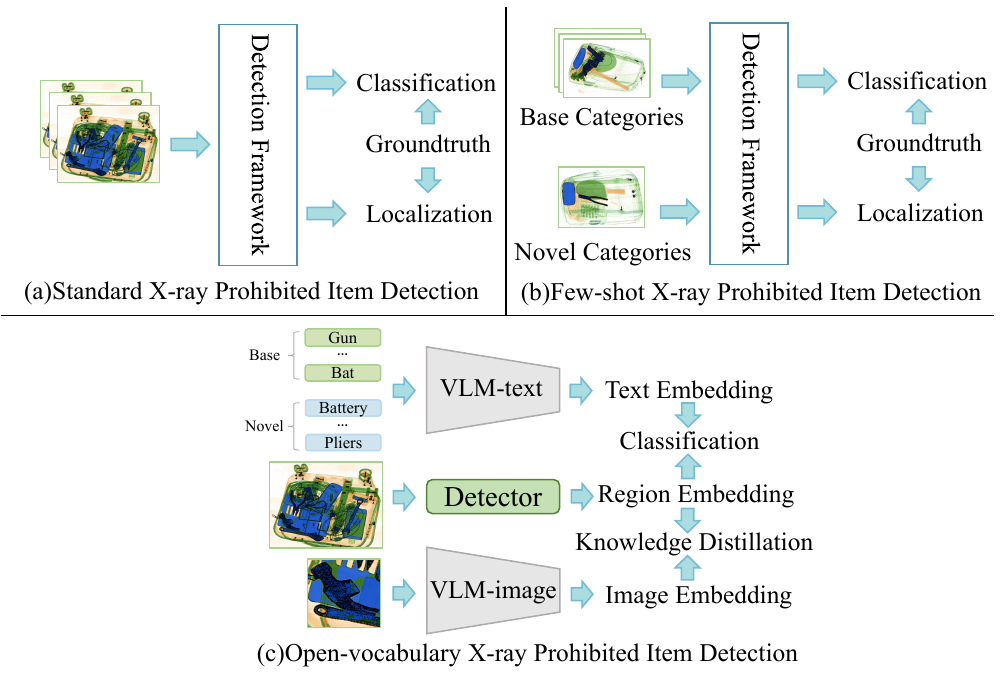}
    \caption{\textbf{(a)} Standard X-ray prohibited item detection requires annotations for all categories during training. \textbf{(b)} Few-shot X-ray prohibited item detection aims to scale the object detector to recognize more categories with only a few samples of novel categories. \textbf{(c)} Open-vocabulary X-ray prohibited item detection uses annotated base categories for training and extends detector to cover unlabeled novel categories which are unknown during the training phase.}
    \label{intro1}
\end{figure}

\IEEEPARstart{S}{ecurity} baggage check system based on X-ray has been widely used in subway stations, train stations, airports, and other public places to maintain public safety. In real security check scenarios, baggage is fed into X-ray security inspection machines, and X-ray images of the baggage are captured by irradiating the objects inside with X-ray and rendering them with pseudo colors according to their spectral absorption rates. Therefore, there is a significant difference between X-ray baggage images and natural images. For example, X-ray baggage images often face severe object occlusion, have a noisy background, as well as lack of texture details and color features, making prohibited item detection challenging. To advance the development of X-ray prohibited item detection, recent efforts have been made to construct security inspection benchmarks. SIXray \cite{x7}, OPIXray \cite{x3}, PIDray \cite{x1}, HiXray \cite{x4}, CLCXray \cite{x11} and PIXray \cite{x2} datasets have been collected while various baselines are established. Although these works have achieved remarkable performance in prohibited item detection, they all fall within the standard close-set object detection paradigm (Fig. \ref{intro1}(a)), i.e., only previously known categories in training are presented during testing, which need costly training and expensive annotations when learning more object categories due to supervision requirements.

Recently, some works attempt to reduce the need for abundant training samples and expensive annotations of novel categories via few-shot object detection (FSOD). FSOD uses abundant annotated samples of base categories and few annotated samples of novel categories for training, aiming to achieve comparable detection performance for novel categories (Fig. \ref{intro1}(b)). To benchmark and promote FSOD studies on X-ray scenarios, \cite{x5} contributes X-ray FSOD dataset as well as weak-feature enhancement network. \cite{x6} presents the Xray-PI dataset and proposes a few-shot X-ray prohibited item segmentation method using patch-based self-supervised learning and prototype reverse validation. Although FSOD can detect more object categories from few labeled training samples, it still operates within a close-set detection paradigm. 

\begin{figure}[H]
    \centering
    \includegraphics[height=5.8cm,width=8.8cm]{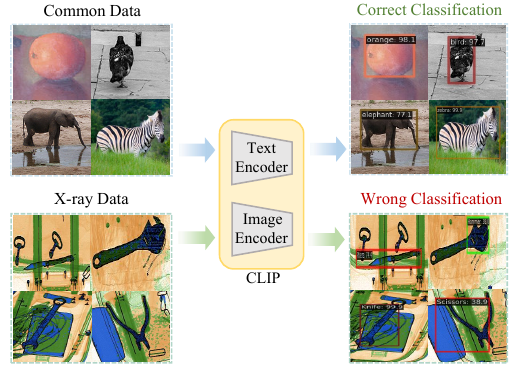}
    \caption{Comparison of CLIP classification results for
    common data in a natural scenario and X-ray data in a security inspection scenario reveals a severe domain shift that leads to failure cases of CLIP.}
    \label{intro2}
\end{figure}

  Considering the fact that in real-world prohibited item detection scenarios, categories of prohibited item are continuously increasing in accordance with the latest laws, developing detection of unknown novel categories is of great significance in X-ray domain. The success of vision-Language Models (VLMs) like contrastive language-image pretraining (CLIP \cite{CLIP}) provides a promising solution to this open-set detection problem. Relying on large-scale image-text pairs for pre-training, CLIP has achieved impressive results on image classification in both zero-shot and transfer learning settings. However, directly applying CLIP to X-ray domain still face challenges due to domain shift between general domain data used for pre-training CLIP and X-ray data. As illustrated in Fig. \ref{intro2}, a pre-trained CLIP model fails to classify prohibited item in X-ray baggage images while succeeding in common data classification. Therefore, detector suffers from a major drop in accuracy when directly using CLIP in X-ray security inspection scenario. To address aforementioned issue and meet the need for detecting novel prohibited item categories that are unknown in advance, we introduce knowledge distillation-based open-vocabulary object detection (OVOD) task to X-ray security inspection domain by adapting CLIP to X-ray domain (Fig. \ref{intro1}(c)).

Specifically, we propose learnable X-ray feature adapter and apply it to CLIP to precisely integrate knowledge from X-ray security inspection domain with the knowledge learned by the pre-trained CLIP thus overcome the sharp performance drop caused by domain shift. X-ray feature adapter consists of three adapter submodules named X-ray Space Adapter (XSA), X-ray Aggregation Adapter (XAA) and X-ray Image Adapter (XIA) respectively. All adapter submodules adopt a bottleneck structure which includes a down-projection linear layer, a hidden linear layer, an up-projection linear and non-linear activation. To leverage the functionality of three adapter submodules, we apply them to different positions within CLIP. XSA and XAA are placed within each standard vision transformer (ViT \cite{ViT}) block of CLIP text encoder while XIA is placed within each standard ViT block in CLIP image encoder. Besides, we keep ViT blocks of CLIP frozen except for the last block of the text and image encoder to leverage extensive knowledge in pre-trained CLIP while incorporating new knowledge in the X-ray domain. By applying X-ray feature adapter and fine-tuning CLIP we form a novel \textbf{O}pen-\textbf{V}ocabulary \textbf{X}-ray prohibited item \textbf{D}etection (OVXD) model.

We conduct extensive experiments on two mainstream public X-ray prohibited item datasets, PIXray and PIDray. First, we validate that OVXD outperforms existing baseline methods \cite{o7,o5,o1}, by achieving 21.0 (15.2 increase) box AP$_{50}$ and 26.0 (17.0 increase) box AP$_{25}$ of novel categories on PIXray while 27.8 (1.5 increase) box AP$_{50}$ and 36.1 (2.7 increase) box AP$_{25}$ of novel categories on PIDray. Second, we prove the effectiveness of the proposed X-ray feature adapter in X-ray scenario compared to previous adapters. Third, we conduct abundant ablation studies. Finally, we verify the generalization ability of OVXD across several X-ray prohibited item datasets.

We summarize main contributions of this work as follows:
\begin{itemize}
    \item[$\bullet$] We extend X-ray prohibited item detection task from a close-set to an open-set by introducing open-vocabulary object detection task to the X-ray domain. To the best of our knowledge, this paper is the first work to study the open-vocabulary object detection method in X-ray security inspection scenario.
    \item[$\bullet$] To effectively address the domain shift that severely impacts detection performance, we propose X-ray feature adapter which adapts CLIP to X-ray OVOD task by integrating domain-specific knowledge with original learned knowledge.
    \item[$\bullet$] We conduct extensive experiments on two widely used X-ray datasets, PIXray and PIDray. The experiment results demonstrate OVXD consistently outperforms baseline methods.
\end{itemize}

The rest of this paper is organized as follows. Section \ref{sec:related work} provides a brief review of related works. Section \ref{sec:methods} gives detailed descriptions of our proposed X-ray feature adapter and its application within CLIP in OVOD framework. Section \ref{sec:experiments} presents experiments and visualizations. Finally, Section \ref{sec:conclusion} concludes the paper.

\section{Related Work}
\label{sec:related work}
\subsection{X-ray Prohibited Item Detection}
X-ray prohibited item detection \cite{x8,x9,x10} not only faces the challenges present in natural images, e.g., scale and intra-class variance, but also suffers from severe object overlapping and color/texture fading in X-ray baggage images, making it a challenging task. Many researchers have devoted their efforts to X-ray prohibited item detection \cite{x12,x13} and several X-ray prohibited item datasets have been collected as well. \cite{x7} presented a large-scale Security Inspection X-ray (SIXray) dataset and utilized a class-balanced hierarchical refinement method to address the challenge of overlapping X-ray baggage images. \cite{x3} contributed Occluded Prohibited Items X-ray (OPIXray) dataset and proposed a de-occlusion attention module to enhance the performance of detectors on occluded prohibited item detection. In \cite{x1}, PIDray dataset was proposed towards real-world prohibited item detection and a selective dense attention network was designed for detecting deliberately hidden items. \cite{x4} presented High-quality X-ray (HiXray) dataset and a lateral inhibition module was proposed inspired by the recognition process of human. \cite{x11} released Cutters and Liquid Containers X-ray (CLCXray) dataset and proposed a new label-aware mechanism to address the overlap problem. \cite{x2} presented Prohibited Item Xray (PIXray) dataset and dense de-overlap attention snake to overcome severe object overlapping. However, different from our paradigm, all these methods are designed for close-set detection, meaning the detector can only detect objects belonging to known categories used for training while failing to detect objects of unknown categories.
\subsection{Open-Vocabulary Object Detection}
As a general and practical paradigm, OVOD \cite{o8,o9,o10,o11} aims at detecting novel object categories that are not seen during detector training. The concept of OVOD was initially put forward by OVR-CNN \cite{o6}, which utilized low-cost image-caption pairs to learn a visual-semantic space and built a two-stage training framework. With recent advancement of VLMs, ViLD \cite{o3} employed VLMs such as CLIP \cite{CLIP} in OVOD and directly distilled vision and language knowledge from pre-trained CLIP into a two-stage detector. Building upon DETR \cite{detr}, OV-DETR \cite{o7} utilized text or exemplar image embeddings from CLIP as conditional queries and matched them with corresponding boxes. RegionCLIP \cite{o5} adopted a two-stage pre-training mechanism to adapt CLIP to encode region-level visual features, and enabling fine-grained alignment between image regions and textual concepts. Similarly, CORA \cite{o4} employed a DETR-style framework to adapt CLIP for OVOD through region prompting and anchor pre-matching. BARON \cite{o1} pointed out that the image-text pairs used for pre-training VLMs usually contain a bag of semantic concepts and proposed a method to align the embedding in bag of different regions rather than only individual regions. While the excellent works described above have achieved advanced performance, they are based on natural image benchmarks, e.g., COCO \cite{coco} and LVIS \cite{lvis}. In contrast, we conducting open-vocabulary object detection in X-ray security inspection scenario.
\subsection{Adapter-based Fine-tuning Adaptation Technique}
VLMs \cite{ALIGN,oscar,vilbert} such as CLIP \cite{CLIP} demonstrate impressive zero-shot classification performance in general domains. However, CLIP still require fine-tuning in fine-grained domains (e.g., remote sensing and medical imaging) due to domain shift and differences in data distribution. To bridge the domain gap, many recent studies have focused on improving the generalization performance of CLIP through adapter-based fine-tuning, instead of training all CLIP backbone parameters (i.e., full fine-tuning). These methods proposed additional learnable adapter modules and parameters of pre-trained CLIP are frozen while parameters in adapter modules are learnable during training on downstream tasks. For example, CLIP-adapter \cite{a1} added feature adapters on top of either visual or language branch of CLIP to incorporate additional knowledge. Following a similar architecture, TIP-adapter \cite{a3} learned additional parameters in a training-free manner by creating weights via a key-value cache model constructed from the few-shot training set. Besides, adapter-based approach has been applied to adapt pre-trained ViTs \cite{ViT} to scalable downstream tasks. AdaptFormer \cite{a5} equipped the transformer encoder with a lightweight adapter module called AdaptMLP. \cite{a2} extended the capabilities of the SAM model to the medical domain by deploying the proposed Medical SAM Adapter for each ViT block in the image encoder. Different from aforementioned works, our X-ray feature adapter focuses on adapting CLIP to OVOD downstream task in X-ray security inspection scenario.

\section{Method}
\label{sec:methods}
In this section, we elaborate on the details of proposed X-ray feature adapter and its application within CLIP in OVOD framework to develop \textbf{O}pen-\textbf{V}ocabulary \textbf{X}-ray prohibited item \textbf{D}etection network, i.e., OVXD model, which marks the first attempt to conduct OVOD task in the X-ray security inspection scenario. An illustration of OVXD is shown in Fig. \ref{NetworkArchitecture}.
\subsection{Preliminaries}
\begin{figure}[h]
    \centering  
    \includegraphics[height=3.7cm,width=2.7cm]{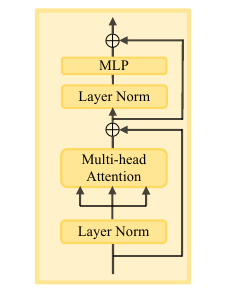}
    \caption{Architecture of vanilla ViT block.}
    \label{ViT}
\end{figure}
\noindent \textbf{CLIP Architecture}. To extract features from vision and language modalities, CLIP includes a image encoder $\mathcal{I}(\cdot)$ and a text encoder $\mathcal{T}(\cdot)$. Both $\mathcal{I}(\cdot)$ and $\mathcal{T}(\cdot)$ adopt a 12-layer transformer, in which each ViT \cite{ViT} block consists of two sub-layers, i.e., multi-head self-attention (MHSA) and multi-layer perceptron (MLP) layer. An illustration is shown in Fig. \ref{ViT}, assuming x$_{tok}$ denotes input tokens, the output x$_{out}$ of the ViT block can be formulated as
\begin{equation}
\begin{aligned}
    x_{att} =MHSA(LN(x_{tok}))+x_{tok},
\end{aligned}
\end{equation}
\begin{equation}
    x_{out} =MLP(LN(x_{att})+x_{att}).
\end{equation}

where LN($\cdot$) denotes layer norm \cite{layernorm} and x$_{att}$ are the tokens produced by MHSA.

\noindent \textbf{Task Definition}. In OVOD setting, only predetermined base categories $\mathcal{C}^{base}$ are annotated and used for training, the object detector is required to generalize beyond $\mathcal{C}^{base}$ to detect novel categories $\mathcal{C}^{novel}$ which is unknown in advance by simply switching the classifier to the novel class name embedding during the test phase. The aim of distillation-based OVOD methods is to obtain an open-vocabulary detector which incorporates $\mathcal{C}^{novel}$ by distilling the knowledge from VLMs like CLIP into a close-set detector trained on $\mathcal{C}^{base}$. In order to detect $\mathcal{C}^{novel}$, previous distillation-based OVOD methods directly use frozen CLIP. However, directly applying CLIP to X-ray open-vocabulary security inspection scenario leads to poor performance due to the domain shift. To further exploit the power of CLIP in X-ray domain, we transition from using a fully frozen CLIP to fine-tuning CLIP with trainable parameters in X-ray feature adapter.
\subsection{The Architecture of X-ray Feature Adapter}
Since existing distillation-based OVOD methods have failed to achieve satisfactory performance in the X-ray security inspection scenario due to the domain shift, X-ray feature adapter is designed to adapt CLIP to open-vocabulary X-ray prohibited item detection task by bridging the gap between X-ray data and general data used for pre-training CLIP, as well as facing the dilemma of heavy occlusion and color/texture fading in X-ray baggage images. 

\begin{figure}[h]
\centering
\includegraphics[width=0.9\linewidth]{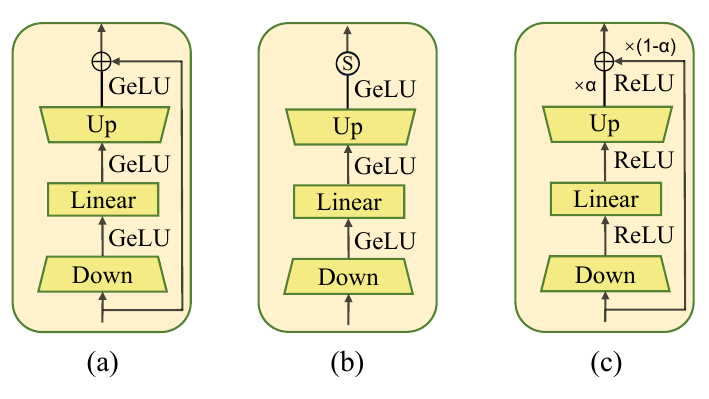}
\caption{Three core submodules of X-ray feature adapter used in paper. Adapter submodules are of bottleneck architecture which consist of a down-projection linear layer, a hidden linear layer and an up-projection linear layer. \textbf{(a)} X-ray Space Adapter. \textbf{(b)} X-ray Aggregation Adapter. \textbf{(c)} X-ray Image Adapter. \textbf{s} represents scale factor.}
\label{adapter}
\end {figure}
As illustrated in Fig. \ref{adapter}, X-ray feature adapter consists of three adapter submodules, i.e., X-ray Space Adapter (XSA), X-ray Aggregation Adapter (XAA) and X-ray Image Adapter (XIA). The design principle of X-ray feature adapter is simple yet effective. All adapter submodules adopt a bottleneck structure that sequentially use a down-projection linear layer with parameters $W_d$, a hidden linear layer with parameters $W_h$, and an up-projection linear layer with parameters $W_u$, which can limit the number of parameters. The down-projection reduces given embedding to a smaller dimension using a simple linear layer. The hidden linear layer effectively integrates more specific knowledge of fine-grained X-ray domain to address challenges posed by domain gap, heavy occlusion and color/texture fading, which significantly benefits the X-ray adaptation capability of CLIP. The up-projection expands reduced embedding back to its original dimension using another linear layer. In addition, there is a non-linear activation layer between each two linear layers for non-linear property. 

Although these adapter submodules share the same bottleneck structure, they exhibit some structural differences. In XSA, a residual connection is adpoted to avoid forgetting original knowledge in CLIP and GeLU is used as non-linear activation. In XAA, we set a constant scale factor $s$ to balance the task-agnostic features generated by the original CLIP and the task-specific features generated by XAA. Similarly, GeLU is used as non-linear activation. XIA also adopts a residual connection to integrate X-ray domain-specific knowledge with general domain knowledge and $\alpha$ is used to balance the ratio between newly learned knowledge and original knowledge of CLIP. ReLU activation is chosen as non-linear activation in XIA.
For specific input embedding $e_1$, $e_2$ and $e_3$, letting $A_S(\cdot)$, $A_A(\cdot)$ and $A_I(\cdot)$ denote XSA, XAA and XIA respectively, three submodules of X-ray feature adapter can be summarized as
\begin{equation} 
    A_S(e_1) = GeLU(GeLU(GeLU(e_1W_d^S)W_h^S)W_u^S)+e_1.
\end{equation}
\begin{equation} 
    A_A(e_2) = s \cdot GeLU(GeLU(GeLU(e_2W_d^A)W_h^A)W_u^A).
\end{equation}
\begin{equation}
    e^* = ReLU(ReLU(ReLU(e_3W_d^I)W_h^I)W_u^I),
\end{equation}
\begin{equation}
    A_I(e_3) = \alpha \cdot e^* + (1-\alpha) \cdot e_3.
\end{equation}

Since manually choosing the hyperparameter $\alpha$ is time-consuming and laborious, we set it as a learnable parameter that can be dynamically optimized from visual features in an adaptive manner. Specially, $\alpha$ is configured with zero initialization and gradually learns to assign a weight between 0 and 1. It tends to be a larger value, approaching 1, when the object's  appearance in X-ray security images is quite different from that in natural images, and tends to be a smaller value, approaching 0, when the object appearance is similar in X-ray security images and natural images. In this way, the features of X-ray prohibited item can be optimally learned and integrated, which enables CLIP to simultaneously leverage its powerful pre-trained knowledge and newly acquired knowledge in X-ray domain, thereby enhancing its ability to detect novel prohibited item categories. 

\subsection{Applying X-ray Feature Adapter to CLIP}
Our starting point is to enhance the X-ray security check capability of OVOD by adapting CLIP to our specific X-ray domain. To bridge the distribution gap between the general data used for pre-training CLIP and the fine-grained data of the X-ray domain, the motivation is to add domain-specific knowledge without forgetting original knowledge. To achieve this goal, we first unfreeze the last ViT block of both $\mathcal{I}(\cdot)$ and $\mathcal{T}(\cdot)$ of CLIP while keeping the other 11 ViT blocks frozen, instead of fully turning all parameters. The reasons behind this are twofold: on the one hand, due to over-parameterization of CLIP and lack of enough training samples in X-ray prohibited item datasets, tuning more parameters would lead to overfitting and slow training process due to forward and backward propagation across more CLIP layers. On the other hand, unfreezing more ViT blocks would harm original knowledge of pre-trained CLIP and lead to a decline in its zero-shot classification performance.

To further integrate X-ray domain-specific knowledge and improve the detection of novel prohibited item categories, we apply three core and efficient submodules of X-ray feature adapter to specific positions within $\mathcal{T}(\cdot)$ and $\mathcal{I}(\cdot)$ inspired by \cite{a1,a2,a5}. In $\mathcal{T}(\cdot)$, we deploy two submodules, XSA and XAA, within each standard ViT block. Specifically, XSA is applied after MHSA and before residual connection while XAA is applied in the residual path of the MLP layer after layer norm to adapt the MLP-enhanced embedding(left side of Fig. \ref{NetworkArchitecture}). In $\mathcal{I}(\cdot)$, we deploy XIA in the residual path of layer norm and MLP layer within each standard ViT block (right side of Fig. \ref{NetworkArchitecture}). In this way, adapter-equipped CLIP is obtained. We denote $\mathcal{T}(\cdot)$ equipped with XSA and XAA as $\mathcal{AT}(\cdot)$, $\mathcal{I}(\cdot)$ equipped with XIA as $\mathcal{AI}(\cdot)$. Note that only the parameters within X-ray feature adapter and the last ViT block of $\mathcal{AI}(\cdot)$ and $\mathcal{AT}(\cdot)$ are trainable while other parameters in CLIP kept frozen.  

\subsection{Applying Adapter-equipped CLIP to OVXD}
\begin{figure*}[ht]
\centering
\includegraphics[width=1.\linewidth]{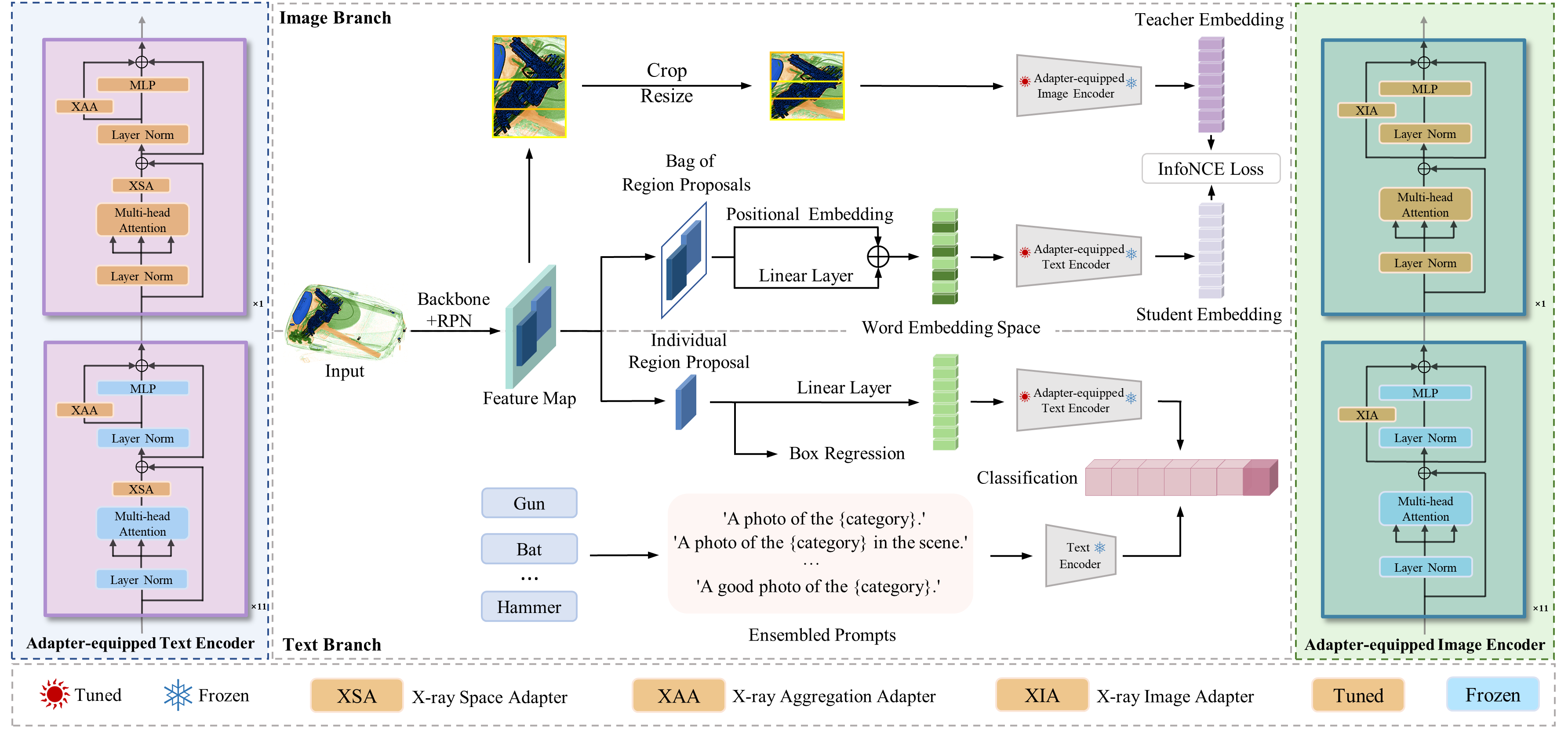}
\caption{An overview of OVXD model. OVXD consists of text and image branch, and open-vocabulary detector is a Faster R-CNN whose classifier is replaced by a linear layer to map region features into word embedding space. OVXD is implemented by applying X-ray feature adapter within CLIP in OVOD framework. Three core adapter submodules of X-ray feature adapter are applied to different positions of the ViT blocks in text and image encoder of CLIP to integrate domain-specific knowledge with original knowledge.}
\label{NetworkArchitecture}
\end{figure*}
As shown in Fig. \ref{NetworkArchitecture}, by using adapter-equipped CLIP we form OVXD model which mainly include two branches: the text branch and the image branch.

In the text branch, since the open-vocabulary classification model is trained on full sentences, category names are filled into a hand-crafted prompt template, e.g., \verb|'A photo of the {category} in the scene'|, and then fed into the frozen $\mathcal{T}(\cdot)$ of CLIP to obtain pre-computed class name embeddings. Following ViLD \cite{o3}, an ensemble of multiple prompt templates is used. In order to detect objects from a wide range of vocabularies, a standard detector, Faster-RCNN \cite{fasterrcnn}, is modified by replacing the original classifier with a linear layer that projects each region feature into the word embedding space to obtain word embedding. The word embedding is further fed into $\mathcal{AT}(\cdot)$, and then the cosine similarity with all class name embeddings is calculated to obtain the final classification results. Given $N$ object categories and the corresponding class name embedding {\it f$_n$}, for a region proposal and its word embedding e$_w$, we adopt the common practice to calculate the probability of the region being classified as the $n$-th category, which can be expressed as
\begin{equation}
\label{equ1}
    logit_n = \frac{\exp(sim\langle\mathcal{AT}(e_w),f_n\rangle\cdot\tau)}{\sum_{i=0}^{\it{N}-1}\exp(sim\langle\mathcal{AT}(e_w),f_i\rangle\cdot\tau)} .
\end{equation}

where $sim\langle\mathcal{AT}(e_w),f\rangle=\mathcal{AT}(e_w)^\top f/(\|\mathcal{AT}(e_w)\|\|f\|)$ and $\tau$ is the temperature to re-scale the value. 

In the image branch, after generating proposals by region proposal network (RPN) \cite{fasterrcnn}, we adopt the similar insight of \cite{o1} to group contextually interrelated region proposals into a bag to help the model better understand the co-occurrence of visual and semantic concepts in the scene. The features of each region proposal in bag are then projected into word embedding space as word embedding e$_{w}$ through a linear layer. The bag-of-regions embedding is obtained by concatenating each word embedding. To retain the spatial information of each region proposal in bag, the shape together with the center position of each proposal are encoded into positional embedding e$_{pos}$ and added to e$_{w}$ before concatenation. Finally student embedding e$_{stu}$ is obtained by encoding bag-of-regions embedding with $\mathcal{AT}(\cdot)$. Letting $k$ denotes group size, the process of forming e$_{stu}$ can be summaried as
\begin{equation}
    e_{stu} =\mathcal{AT}(e_w^0+e_{pos}^0,e_w^1+e_{pos}^1,...,e_w^{k-1}+e_{pos}^{k-1}).
\end{equation}

Meanwhile, image crops enclosing grouped region proposals are sent to $\mathcal{AI}(\cdot)$ to generate image embeddings (i.e., teacher embedding). Chosen as the teacher, $\mathcal{AI}(\cdot)$ is expected to distill information to the student open-vocabulary detector by aligning student and teacher embeddings via InfoNCE loss \cite{InfoNCEloss}.

OVXD is an end-to-end framework which does not require pre-training on CLIP using X-ray security inspection images. In the training phase, the learning on base categories follows standard loss function of Faster-RCNN for classification and regression: 
\begin{equation}
    \mathcal L_{total} = 
    \mathcal L_{rpn}^{cls} + L_{rpn}^{bbox} +\mathcal L_{cls} + \mathcal L_{bbox}.
\end{equation}

where $\mathcal L_{rpn}^{cls}$ is classification loss in RPN \cite{fasterrcnn} to estimate the probability that if an anchor contains object and $\mathcal L_{rpn}^{bbox}$ is regression loss to refine the anchor, $\mathcal L_{cls}$ is cross-entropy loss for classifying the object and $\mathcal L_{bbox}$ is smooth L1 loss for bounding box regression.
\begin{table*}[t]
\renewcommand{\arraystretch}{1.3}
\caption{AP50 and AP25 results comparison between OVXD and baseline open-vocabulary object detection methods on PIXray dataset. Batter, pliers, scissors and wrench are novel categories.}
\label{tab1}
\tabcolsep=0.017\linewidth
\setlength\tabcolsep{4.1pt}
\begin{tabular*}{\hsize}{c|c|ccccccc|ccccccc}
\hline
\multirow{2}{*}{Method} &
\multirow{2}{*}{Venue} &
\multicolumn{7}{c|}{\multirow{1}{*}{AP$_{50}$}} & 
\multicolumn{7}{c}{\multirow{1}{*}{AP$_{25}$}} 
\\
&&Battery&Pliers&Scissors&Wrench&Base&Novel&All&Battery&Pliers&Scissors&Wrench&Base&Novel&All\\
\hline
OV-DETR \cite{o7}&ECCV2022&0.0&0.1&0.2&0.1&65.2&0.1&47.8&0.0&0.6&0.9&0.2&66.1&0.4&48.6\\
RegionCLIP \cite{o5}&CVPR2022&0.0&2.4&0.4&2.6&78.1&1.3&57.7&0.0&8.4&3.1&10.4&85.7&5.5&64.3\\
BARON \cite{o1}&CVPR2023&0.6&3.1&1.5&18.1&81.0&5.8&60.9&1.1&6.2&5.2&23.7&83.2&9.0&63.4\\
\hline
OVXD (Ours)
&\text{-}&\textbf{14.6}&\textbf{6.4}&\textbf{16.9}&\textbf{46.2}&\textbf{85.9}&\textbf{21.0}&\textbf{68.6}&\textbf{17.4}&\textbf{12.4}&\textbf{22.8}&\textbf{51.3}&\textbf{87.3}&\textbf{26.0}&\textbf{71.0}\\
Improvement&\text{-}&\textbf{\textcolor[RGB]{209,41,32}{+14.0}}&\textbf{\textcolor[RGB]{209,41,32}{+3.3}}&\textbf{\textcolor[RGB]{209,41,32}{+15.4}}&\textbf{\textcolor[RGB]{209,41,32}{+28.1}}&\textbf{\textcolor[RGB]{209,41,32}{+4.9}}&\textbf{\textcolor[RGB]{209,41,32}{+15.2}}&\textbf{\textcolor[RGB]{209,41,32}{+7.7}}&\textbf{\textcolor[RGB]{209,41,32}{+16.3}}&\textbf{\textcolor[RGB]{209,41,32}{+6.2}}&\textbf{\textcolor[RGB]{209,41,32}{+17.6}}&\textbf{\textcolor[RGB]{209,41,32}{+27.6}}&\textbf{\textcolor[RGB]{209,41,32}{+4.1}}&\textbf{\textcolor[RGB]{209,41,32}{+17.0}}&\textbf{\textcolor[RGB]{209,41,32}{+7.6}}\\
\hline
\end{tabular*}
\end{table*}
\begin{table*}[t]
\renewcommand{\arraystretch}{1.3}
\caption{AP50 and AP25 results comparison between OVXD and baseline open-vocabulary object detection methods on PIDray dataset. Scissors, wrench and handcuffs are novel categories.}
\label{tab2}
\tabcolsep=0.017\linewidth
\setlength\tabcolsep{5.4pt}
\begin{tabular*}{\hsize}{c|c|cccccc|cccccc}
\hline
\multirow{2}{*}{Method} &
\multirow{2}{*}{Venue} &
\multicolumn{6}{c|}{\multirow{1}{*}{AP$_{50}$}} & 
\multicolumn{6}{c}{\multirow{1}{*}{AP$_{25}$}}
\\
&&Scissors&Wrench&Handcuffs&Base&Novel&All&Scissors&Wrench&Handcuffs&Base&Novel&All\\
\hline
OV-DETR \cite{o7}&ECCV2022&1.2&0.3&0.4&0.5&0.6&0.5&4.3&1.4&0.8&2.1&2.2&2.2\\
RegionCLIP \cite{o5}&CVPR2022&4.2&7.0&0.1&33.6&3.8&26.2&11.7&12.8&1.5&\textbf{45.1}&8.7&36.0\\
BARON \cite{o1}&CVPR2023&13.1&21.6&44.3&26.7&26.3&26.6&16.4&27.5&56.3&29.8&33.4&30.7\\
\hline
OVXD (Ours)&\text{-}&\textbf{14.7}&\textbf{23.4}&\textbf{45.3}&\textbf{33.7}&\textbf{27.8}&\textbf{32.2}&\textbf{19.8}&\textbf{27.7}&\textbf{60.7}&37.7&\textbf{36.1}&\textbf{37.3}\\
Improvement&\text{-}&\textbf{\textcolor[RGB]{209,41,32}{+1.6}}&\textbf{\textcolor[RGB]{209,41,32}{+1.8}}&\textbf{\textcolor[RGB]{209,41,32}{+1.0}}&\textbf{\textcolor[RGB]{209,41,32}{+7.0}}&\textbf{\textcolor[RGB]{209,41,32}{+1.5}}&\textbf{\textcolor[RGB]{209,41,32}{+5.6}}&\textbf{\textcolor[RGB]{209,41,32}{+3.4}}&\textbf{\textcolor[RGB]{209,41,32}{+0.2}}&\textbf{\textcolor[RGB]{209,41,32}{+4.4}}&\textbf{\textcolor[RGB]{209,41,32}{-7.4}}&\textbf{\textcolor[RGB]{209,41,32}{+2.7}}&\textbf{\textcolor[RGB]{209,41,32}{+6.6}}\\
\hline
\end{tabular*}
\end{table*}
\subsection{Difference between X-ray Feature Adapter and Previous Adapters}
\begin{figure}[H]
    \centering
    \includegraphics[height=5.8cm,width=8.8cm]{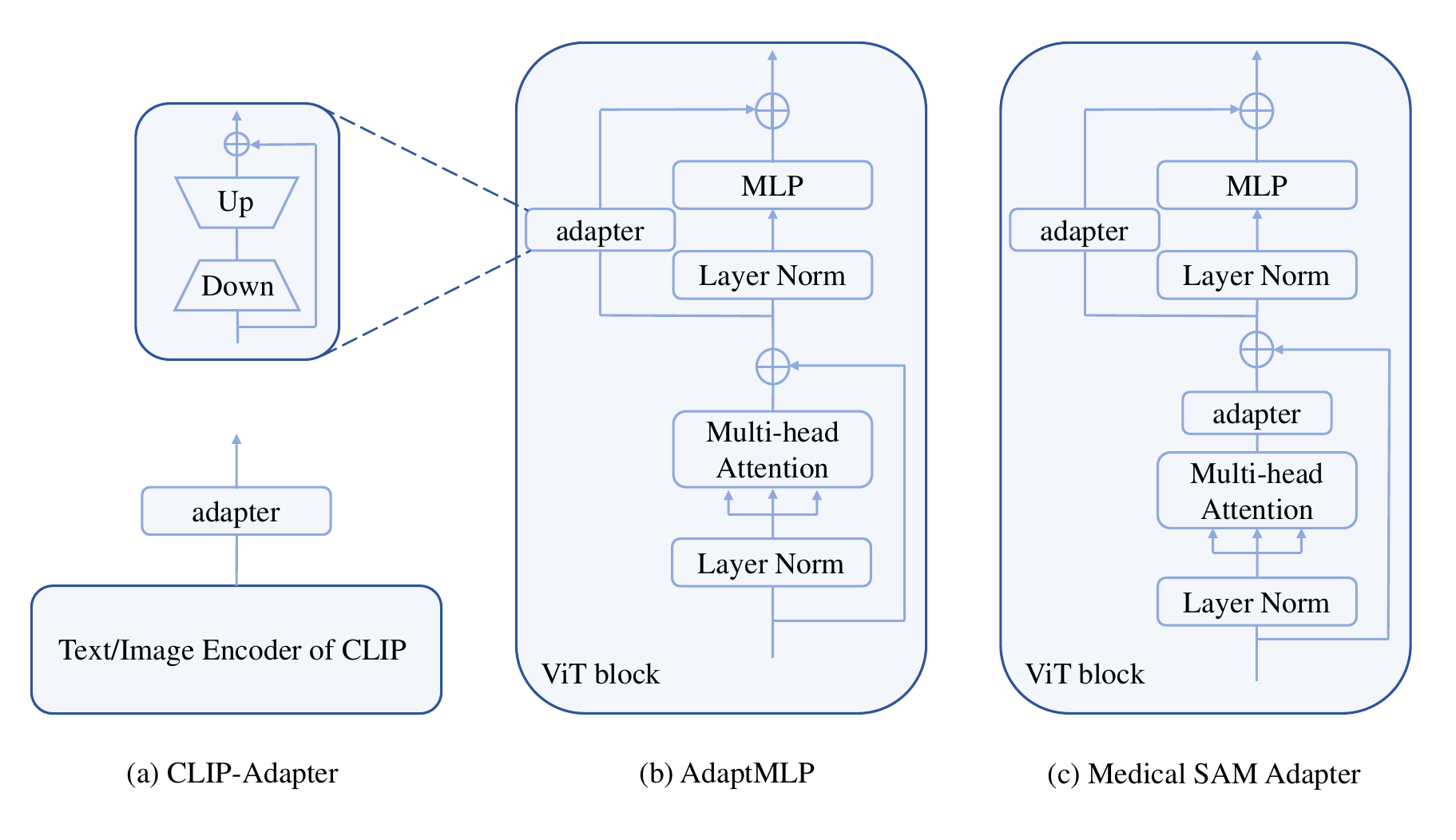}
    \caption{The positions and architectures of previous adapters.}
    \label{compare_adapter}
\end{figure}

Compared with CLIP-Adapter\cite{a1}, AdaptMLP \cite{a5} and Medical SAM Adapter \cite{a2} which are shown in Fig. \ref{compare_adapter}, X-ray feature adapter appears two significant differences: $(1)$ X-ray feature adapter combines three adapter submodules together and applies them to different positions within both $\mathcal{T}(\cdot)$ and $\mathcal{I}(\cdot)$. $(2)$  X-ray feature adapter improves the architecture of existing adapters. 

Regarding the positions of adapter, CLIP-Adapter places adapter on the top of text branch or image branch (Fig. \ref{compare_adapter}(a)) while AdaptMLP places adapter only in the residual path of layer norm and MLP layer in each standard ViT block (Fig. \ref{compare_adapter}(b)). Medical SAM Adapter applies two adapters in each standard ViT blocks, one is after MHSA and before residual connection, the other one is in the residual path of layer norm and MLP layer (Fig. \ref{compare_adapter}(c)). However, we find that directly applying adapter in the same positions as in previous works yields unsatisfactory performance. Thus we propose X-ray feature adapter which uses a mix of three submodules at different positions. Such design makes X-ray feature adapter more flexible and general to visual and textual modality. Detailed experiment results are shown in Table \ref{tab3}.

As for architecture, CLIP-Adapter, AdaptMLP and Medical SAM Adapter all adopt the same architecture of adapter which sequentially uses a down-projection and an up-projection with a residual. To learn more knowledge specific to the X-ray security inspection domain, we add an hidden linear layer between down-projection linear layer and up-projection linear layer. With more specific features learned, the model can better handle heavy occlusion and color/texture fading, which further enhancing the performance of detecting novel categories.

\section{Experiments}
\label{sec:experiments}
In this section, we evaluate the proposed OVXD by conducting extensive experiments. \textbf{First}, we describe our experimental settings, including datasets, evaluation metrics and implementation details. \textbf{Second}, we compare our OVXD with other baseline OVOD methods to demonstrate its superiority and provide a comprehensive analysis. \textbf{Third}, we compare X-ray feature adapter with previous adapters to verify its effectiveness in X-ray scenario. \textbf{Fourth}, we perform ablation studies to prove the effectiveness of each adapter submodule as well as explore different experimental configurations. \textbf{Finally}, we directly transfer trained OVXD to different X-ray prohibited item datasets to certify its generalization ability.
\begin{table*}[t]
\renewcommand{\arraystretch}{1.3}
\caption{Ablation experiments with different combination of XAA, XSA and XIA submodules on PIXray.}
\label{tab4}
\tabcolsep=0.017\linewidth
\setlength\tabcolsep{5.5pt}
\begin{tabular*}{\hsize}{ccc|ccccccc|ccccccc}
\hline
\multirow{2}{*}{XAA} &
\multirow{2}{*}{XSA} &
\multirow{2}{*}{XIA} &
\multicolumn{7}{c|}{\multirow{1}{*}{AP$_{50}$}} & 
\multicolumn{7}{c}{\multirow{1}{*}{AP$_{25}$}} \\
&&&\textcolor[RGB]{127,127,127}{Battery}&\textcolor[RGB]{127,127,127}{Pliers}&\textcolor[RGB]{127,127,127}{Scissors}&\textcolor[RGB]{127,127,127}{Wrench}&\textcolor[RGB]{127,127,127}{Base}&Novel&\textcolor[RGB]{127,127,127}{All}&\textcolor[RGB]{127,127,127}{Battery}&\textcolor[RGB]{127,127,127}{Pliers}&\textcolor[RGB]{127,127,127}{Scissors}&\textcolor[RGB]{127,127,127}{Wrench}&\textcolor[RGB]{127,127,127}{Base}&Novel&\textcolor[RGB]{127,127,127}{All}\\
\hline
&&&\textcolor[RGB]{127,127,127}{4.2}&\textcolor[RGB]{127,127,127}{6.3}&\textcolor[RGB]{127,127,127}{18.0}&\textcolor[RGB]{127,127,127}{36.6}&\textcolor[RGB]{127,127,127}{83.5}&16.3
&\textcolor[RGB]{127,127,127}{65.6}&\textcolor[RGB]{127,127,127}{7.2}&\textcolor[RGB]{127,127,127}{9.2}&\textcolor[RGB]{127,127,127}{23.0}&\textcolor[RGB]{127,127,127}{46.5}&\textcolor[RGB]{127,127,127}{85.2}&21.5
&\textcolor[RGB]
{127,127,127}{68.2}\\
\checkmark&&&\textcolor[RGB]{127,127,127}{5.8}&\textcolor[RGB]{127,127,127}{5.7}&\textcolor[RGB]{127,127,127}{16.2}&\textcolor[RGB]{127,127,127}{38.3}&\textcolor[RGB]{127,127,127}{84.6}&
16.5&\textcolor[RGB]{127,127,127}{66.4}&\textcolor[RGB]{127,127,127}{7.3}&\textcolor[RGB]{127,127,127}{9.0}&\textcolor[RGB]{127,127,127}{21.6}&\textcolor[RGB]{127,127,127}{45.2}&\textcolor[RGB]{127,127,127}{86.3}&
20.8&\textcolor[RGB]
{127,127,127}{68.8}\\
&\checkmark&&\textcolor[RGB]{127,127,127}{12.4}&\textcolor[RGB]{127,127,127}{4.6}&\textcolor[RGB]{127,127,127}{10.4}&\textcolor[RGB]{127,127,127}{34.1}&\textcolor[RGB]{127,127,127}{84.3}&15.4&\textcolor[RGB]{127,127,127}{65.9}&\textcolor[RGB]{127,127,127}{15.1}&\textcolor[RGB]{127,127,127}{8.2}&\textcolor[RGB]{127,127,127}{15.6}&\textcolor[RGB]{127,127,127}{43.7}&\textcolor[RGB]{127,127,127}{86.3}&20.6&\textcolor[RGB]{127,127,127}{68.8}\\
&&\checkmark&\textcolor[RGB]{127,127,127}{4.4}&\textbf{\textcolor[RGB]{127,127,127}{7.6}}&\textcolor[RGB]{127,127,127}{15.7}&\textcolor[RGB]{127,127,127}{32.3}&\textcolor[RGB]{127,127,127}{83.2}&15.0&\textcolor[RGB]{127,127,127}{65.0}&\textcolor[RGB]{127,127,127}{7.5}&\textbf{\textcolor[RGB]{127,127,127}{13.0}}&\textcolor[RGB]{127,127,127}{21.7}&\textcolor[RGB]{127,127,127}{41.0}&\textcolor[RGB]{127,127,127}{85.1}&20.8&\textcolor[RGB]{127,127,127}{68.0}\\
\checkmark&\checkmark&&\textcolor[RGB]{127,127,127}{7.0}&\textcolor[RGB]{127,127,127}{6.7}&\textcolor[RGB]{127,127,127}{9.2}&\textcolor[RGB]{127,127,127}{44.8}&\textcolor[RGB]{127,127,127}{84.2}&16.9&\textcolor[RGB]{127,127,127}{66.3}&\textcolor[RGB]{127,127,127}{9.6}&\textcolor[RGB]{127,127,127}{10.5}&\textcolor[RGB]{127,127,127}{13.7}&\textbf{\textcolor[RGB]{127,127,127}{52.7}}&\textcolor[RGB]{127,127,127}{85.9}&21.6&\textcolor[RGB]{127,127,127}{68.8}\\
&\checkmark&\checkmark&\textcolor[RGB]{127,127,127}{12.3}&\textcolor[RGB]{127,127,127}{7.1}&\textcolor[RGB]{127,127,127}{10.9}&\textcolor[RGB]{127,127,127}{36.5}&\textcolor[RGB]{127,127,127}{82.7}&16.7&\textcolor[RGB]{127,127,127}{65.1}&\textcolor[RGB]{127,127,127}{15.7}&\textcolor[RGB]{127,127,127}{11.8}&\textcolor[RGB]{127,127,127}{15.6}&\textcolor[RGB]{127,127,127}{45.7}&\textcolor[RGB]{127,127,127}{84.5}&22.2&\textcolor[RGB]{127,127,127}{67.9}\\
\checkmark&&\checkmark&\textcolor[RGB]{127,127,127}{5.6}&\textcolor[RGB]{127,127,127}{7.3}&\textbf{\textcolor[RGB]{127,127,127}{18.4}}&\textcolor[RGB]{127,127,127}{41.3}&\textcolor[RGB]{127,127,127}{84.2}&18.1&\textcolor[RGB]{127,127,127}{66.6}&\textcolor[RGB]{127,127,127}{7.7}&\textcolor[RGB]{127,127,127}{12.5}&\textbf{\textcolor[RGB]{127,127,127}{24.8}}&\textcolor[RGB]{127,127,127}{47.6}&\textcolor[RGB]{127,127,127}{86.0}&23.2&\textcolor[RGB]{127,127,127}{69.3}\\
\checkmark&\checkmark&\checkmark&\textbf{\textcolor[RGB]{127,127,127}{14.6}}&\textcolor[RGB]{127,127,127}{6.4}&\textcolor[RGB]{127,127,127}{16.9}&\textbf{\textcolor[RGB]{127,127,127}{46.2}}&\textbf{\textcolor[RGB]{127,127,127}{85.9}}&\textbf{21.0}&\textbf{\textcolor[RGB]{127,127,127}{68.6}}&\textbf{\textcolor[RGB]{127,127,127}{17.4}}&\textcolor[RGB]{127,127,127}{12.4}&\textcolor[RGB]{127,127,127}{22.8}&\textcolor[RGB]{127,127,127}{51.3}&\textbf{\textcolor[RGB]{127,127,127}{87.3}}&\textbf{26.0}&\textbf{\textcolor[RGB]{127,127,127}{71.0}}\\
\hline
\end{tabular*}
\end{table*}
\begin{table*}[t]
    \renewcommand{\arraystretch}{1.2}
	\noindent
	\begin{minipage}{0.3\textwidth}
		\centering
		\caption{Different reduction ratios of bottleneck layer in XIA.}
		\label{tab6}
		\begin{tabular}{c|c|c}
            \hline
            ratio&AP$_{50}^{Novel}$&AP$_{25}^{Novel}$\\
            \hline
            1&17.1&23.6\\
            2&\textbf{21.0}&\textbf{26.0}\\
            4&18.2&23.3\\
            8&17.9&22.1\\
            \hline
            \end{tabular}
	\end{minipage}
        \hfill
        \begin{minipage}{0.3\textwidth}
		\centering
		\caption{Different reduction ratios of bottleneck layer in XSA and XAA.}
		\label{tab7}
		  \begin{tabular}{c|c|c}
            \hline
            ratio&AP$_{50}^{Novel}$&AP$_{25}^{Novel}$\\
            \hline
            1&19.2&23.9\\
            2&14.8&19.8\\
            4&\textbf{21.0}&\textbf{26.0}\\
            8&16.6&22.3\\
            \hline
            \end{tabular}
	\end{minipage}
        \hfill
        \begin{minipage}{0.3\textwidth}
		\centering
		\caption{Ablation study on scale factor of XAA.}
		\label{tab8}
		  \begin{tabular}{c|c|c}
            \hline
            scale&AP$_{50}^{Novel}$&AP$_{25}^{Novel}$\\
            \hline
            0.25&17.6&22.1\\
            0.50&\textbf{21.0}&\textbf{26.0}\\
            0.75&17.2&21.2\\
            1&20.1&24.9\\
            \hline
            \end{tabular}
	\end{minipage}
\end{table*}
\subsection{Experimental Setup}
\subsubsection{Datasets}
We evaluate our method on two mainstream public X-ray prohibited item datasets, i.e., PIXray \cite{x2} and PIDray \cite{x1}. To conduct OVOD in X-ray security inspection scenario, we need to manually split all categories into separate base categories and novel categories. Since PIXray dataset ensures a balanced distribution of numbers and types of prohibited item, we simulate the ratio of base categories to novel categories in COCO \cite{coco} and divide entire 15 object categories into 11 base categories and 4 novel categories. Specifically, \textit{gun}, \textit{knife}, \textit{lighter}, \textit{hammer}, \textit{screwdriver}, \textit{dart}, \textit{bat}, \textit{fireworks}, \textit{saw blade}, \textit{razor blade} and \textit{pressure vessel} are treated as base categories while \textit{battery}, \textit{pliers}, \textit{scissors} and \textit{wrench} treated as novel categories. As for PIDray dataset, we split all 12 object categories into 9 base categories and 3 novel categories and treat \textit{baton}, \textit{bullet}, \textit{gun}, \textit{hammer}, \textit{powerbank}, \textit{knife}, \textit{lighter}, \textit{pliers} and \textit{sprayer} as base categories while \textit{scissors}, \textit{wrench} and \textit{handcuffs} as novel categories.
\subsubsection{Evaluation Metrics}
We follow OVR-CNN \cite{o6} and OV-3DET \cite{o2} to report the box AP at IoU threshold 0.5 and 0.25, denoted as AP$_{50}$ and AP$_{25}$ respectively. For completeness, we follow the settings of evaluate the detection performance on base categories (Base), novel categories (Novel) and all categories (All). To illustrate the effectiveness of the detector on novel categories, we also provide detection results on each novel categories. AP$_{25}$ and AP$_{50}$ of novel categories (denoted as AP$_{25}^{Novel}$ and AP$_{50}^{Novel}$) are the main metrics that evaluate open-vocabulary detection performance on PIXray and PIDray datasets.
\subsubsection{Implementation Details}
In this work, PyTorch toolkit \cite{pytorch} is used to conduct all experiments on two NVIDIA GeForce RTX 3090 GPUs. We develop the detector on MMdetection \cite{mmdetection} using Faster-RCNN \cite{fasterrcnn} with ResNet50-FPN \cite{fpn} architecture and initialize backbone with weights pre-trained by SOCO \cite{soco}. Synchronized Batch Normalization \cite{SyncBN} is employed following ViLD \cite{o3}. CLIP model is based on ViT-B/32 \cite{ViT} and parameters in X-ray feature adapter is trained without initialization. The entire model is optimized by a stochastic gradient descent (SGD) algorithm with initial learning rate of 0.005, momentum of 0.9 and weight decay of 0.000025. We set the batch size as 2 on each GPU. Considering lack of enough training samples on PIXray and PIDray datasets, we shorten the training schedule to prevent overfitting, e.g., 5000 iterations on PIXray and 10000 iterations for PIDray.
\subsection{Compared with Baselines}
Many existing OVOD methods require caption supervision to learn vision-language alignment from image-caption pairs, however, none of public prohibited item dataset has provided image caption annotations. Thus, we compare OVXD with baseline OVOD methods that use CLIP as supervision, including OV-DETR \cite{o7}, RegionCLIP \cite{o5} and BARON \cite{o1} on PIXray and PIDray. The results are summarized in Table \ref{tab1} and Table \ref{tab2}, respectively. We choose baselines which follow strict open-vocabulary pipeline as we do as much as possible for a fair comparison.

As shown in Table \ref{tab1}, our method significantly outperforms baseline methods across all metrics. Compared with BARON, which is the second highest, OVXD particularly +15.2 AP$_{50}^{Novel}$ and +17.0 AP$_{25}^{Novel}$. It is noteworthy that our method can detect novel categories without decreasing the effectiveness of detecting base categories. On the contrary, OVXD achieves much better performance on both novel and base categories. Besides, due to the distribution gap between X-ray data and general data, baselines perform poorly in detecting novel prohibited item categories, and are almost disabled on severely overlapping novel categories, e.g., \textit{battery} and \textit{scissors}. It can be seen that OVXD can not only bridge domain gap, but also substantially improve performance on categories that suffer from severe overlapping.

Regarding experiments on PIDray, since PIDray groups the test set into three subsets (i.e., easy, hard and hidden) according to the difficulty degree of prohibited item detection, we evaluate detection performance on each subset separately and take the average as the final result. As presented in Table \ref{tab2}, OVXD also obtains the best performance on novel categories by achieving 27.8 AP$_{50}^{Novel}$ and 36.1 AP$_{25}^{Novel}$.

\begin{figure*}[t]
    \centering
    \includegraphics[height=10.5cm,width=18.6cm]{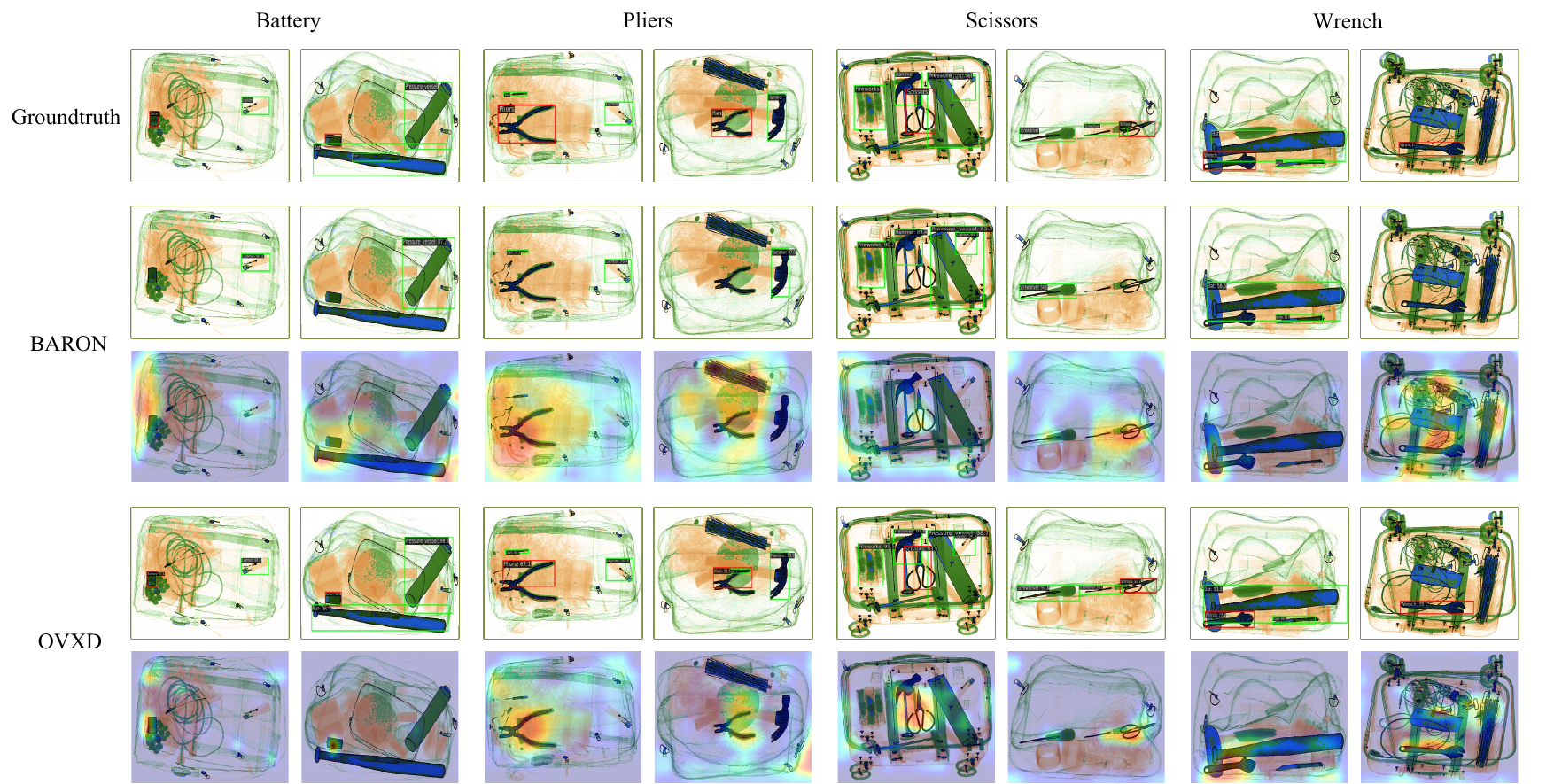}
    \caption{Qualitative comparisons of detection results and attention map visualization between OVXD and BARON. From top to bottom are ground-truth and the results of BARON and OVXD. From left to right are the results of novel categories ‘Battery’, ‘Pliers’, 'Scissors', and ‘Wrench’, respectively. Red boxes are for the novel categories
    while green for the base categories. OVXD successfully detects object of novel categories which are missed by BARON.}
    \label{result}
\end{figure*}
\subsection{Comparison with Previous Adapters}
We designate OVXD without XSA, XAA and XIA as the vanilla model. For fair comparison, we utilize structural variant of CLIP-Adapter \cite{a1} which fine-tunes both the image and text encoders of CLIP model. As for AdaptMLP \cite{a5} and Medical SAM Adapter \cite{a2}, we integrate them within each ViT block in the CLIP image encoder while simultaneously inserting our XSA and XAA in the CLIP text encoder. Experiments are conducted on PIXray dataset and results are displayed in Table \ref{tab3}. Due to placing adapters at inappropriate positions for X-ray security inspection scenario, previous adapters all lead to a decline in AP$_{50}^{Novel}$ to different degrees.
\begin{table}[H]
\caption{Performance comparison between X-ray feature adapters and other three previous adapters.}
\label{tab3}
\resizebox{\linewidth}{!}{
\renewcommand{\arraystretch}{1.2}
    \footnotesize
    \begin{tabular}{c|c|c}
    \hline
    Method&AP$_{50}^{Novel}$&AP$_{25}^{Novel}$\\
    \hline
    vanilla model&16.3&21.5\\
    + CLIP-Adapter \cite{a1}&14.7&25.5\\
    + AdaptMLP \cite{a5}&13.8&24.3\\
    + Medical SAM Adapter \cite{a2}&8.9&17.0\\
    + X-ray feature adapters (Ours)&\textbf{21.0}&\textbf{26.0}\\
    \hline
    \end{tabular}}
\end{table}
\subsection{Ablation Study}
In this section, we conduct ablation studies on PIXray dataset to evaluate the effectiveness of each component in OVXD and explore different configurations of proposed method.
\subsubsection{Effectiveness of Different Adapter Modules}
We list a comprehensive ablation study table detailing the impact of XSA, XAA and XIA. As presented in Table \ref{tab4}, we report how the performance of our method is proved by the addition of different X-ray feature adapter submodules. While the impact of adding a single adapter submodule may not be immediately apparent, a remarkable improvement is observed when two adapter submodules are added at the same time. By simultaneously integrating all three adapter submodules together, our method achieves the highest AP$_{50}$ and AP$_{25}$ on base, novel and all categories, with AP$_{50}^{Novel}$ reaching 21.0\% and AP$_{25}^{Novel}$ reaching 26.0\%. This represents an improvement of 4.7\% and 4.5\%, respectively, compared to the model without X-ray feature adapter, thus confirming the effectiveness of using combination of three adapter submodules.
\subsubsection{Number of Unfrozen Layers in CLIP}
Considering the fact that large distribution gap existing between original data and domain-specific data, alongside the limited training samples for fine-tuning, the most appropriate approach is to unfreeze the last several ViT \cite{ViT} blocks in CLIP backbone. Table \ref{tab5} investigates the impact of the number of unfrozen layers. The number of unfrozen layer ranges from 0 to 6, with the last three layers of both image encoder and text encoder unfrozen at most, and all layers kept frozen at least. It is evident that either unfreezing too much or too few layers significantly degrades performance and the best result for novel categories is achieved when only the last ViT block of both image encoder and text encoder is unfrozen.
\begin{table}[H]
\caption{Ablation study on the number of unfrozen layers. T12, T11, T10 refer to the 12th, 11th and 10th ViT blocks of the text encoder while I12, I11, I10 refer to the 12th, 11th and 10th ViT blocks of the image encoder, respectively. ViT blocks with "\checkmark" are unfrozen.}
\label{tab5}
\resizebox{\linewidth}{!}{
\renewcommand{\arraystretch}{1.2}
    \footnotesize
    \begin{tabular}{cccccc|c|c}
    \hline
    T12&T11&T10&I12&I11&I10&AP$_{50}^{Novel}$&AP$_{25}^{Novel}$\\
    \hline
    &&&&&&13.6&19.1\\
    \checkmark&&&&&&14.6&18.9\\
    &&&\checkmark&&&13.2&18.7\\
    \checkmark&&&\checkmark&&&\textbf{21.0}&\textbf{26.0}\\
    \checkmark&\checkmark&&\checkmark&\checkmark&&12.5&17.1\\
    \checkmark&\checkmark&\checkmark&\checkmark&\checkmark&\checkmark&
    14.9&19.4\\
    \hline
    \end{tabular}}
\end{table}
\subsubsection{Reduction Ratio of Bottleneck Layer}
Intuitively, optimal hidden dimension varies per dataset. Therefore, we conduct ablations to explore optimal value by varying the hidden dimension of bottleneck layers in XSA, XAA and XIA. The results are shown in Table \ref{tab6} and Table \ref{tab7} respectively, where reduction ratio represents the ratio of original feature dimension to hidden feature dimension. By increasing the reduction ratio from 1 to 8, we observe that both too small and too large hidden dimensions significantly deteriorate the performance. The best result on novel categories is obtained with a reduction ratio of 2 in XIA, while it is 4 in XSA and XAA. This configuration allows for the extraction of features from X-ray images without forgetting original semantics.
\subsubsection{Scale Factor of XAA}
To adjust the degree of newly added knowledge for better performance and ensure the stability of fine-tuning, $s$ is empirically set between 0 and 1 for vision tasks. We evaluate multiple scale values, ranging from 0.25 to 1. The results are summarized in Table \ref{tab8}, showing that optimal performance on novel categories can be achieved when $s$ equal to 0.50. Larger or smaller values of $s$ result in different degrees of performance decline.

\subsection{Transfer to Other X-ray Prohibited Item Datasets}
Robustness to data distribution shift is widely used to evaluate the generalization ability of a deep-learning model. Our trained model can be transferred to various X-ray prohibited item datasets by simply switching the classifier to the category text embeddings of new datasets. To evaluate the transfer ability of our proposed OVXD, we transfer the open-vocabulary detector (Faster-RCNN) trained on SIXray \cite{x7} to PIXray and PIDray. Since three datasets have much smaller vocabularies, category overlap is unavoidable. We mainly compare OVXD with BARON, the state-of-the-art (SOTA) method for open vocabulary object detection, and the results are displayed in Table \ref{tab9}. Compared with BARON, our approach exhibits better generalization ability on all datasets. In this way, we can directly transfer a trained detector to different X-ray prohibited item datasets by using language instead of training detector from start, even without further fine-tuning.
\begin{table}[H]
\caption{Comparison of generalization ability of the model trained on SIXray}
\label{tab9}
\resizebox{\linewidth}{!}{
\renewcommand{\arraystretch}{1.2}
    \tiny
    \begin{tabular}{c|cc|cc}
    \hline
    \multirow{2}{*}{Method} &
    \multicolumn{2}{c|}{\multirow{1}{*}{PIXray}} & 
    \multicolumn{2}{c}{\multirow{1}{*}{PIDray}} \\
    &AP$_{25}$&AP$_{50}$&AP$_{25}$&AP$_{50}$\\
    \hline
    BARON \cite{o1}&13.2&11.7&18.5&14.5\\
    ours& \textbf{15.8}&\textbf{14.1}&\textbf{18.8}&\textbf{15.2}\\
    \hline
    \end{tabular}}
\end{table}
\subsection{Visualization Analysis}
We use images from validation set of PIXray for visualization. First, we visualize the detection results of OVXD and SOTA BARON in Fig. \ref{result}. The experimental results show that OVXD successfully detects novel categories (red box) while BARON fails to detect all novel categories. It is noteworthy that OVXD can not only detects novel categories but also improves detection accuracy of base categories (green box).

Furthermore, we present predictions of detectors learned through OVXD and BARON using attention map visualizations. From Fig. \ref{result}, it is evident that the model learned through OVXD generates responses at locations of novel categories and covers object regions more precisely while BARON induces rough, incomplete, or inaccurate responses. To sum up, heatmap visualizations further demonstrate that our proposed method is effective in detecting novel prohibited item categories.

\section{Conclusion}
\label{sec:conclusion}
In this paper, we bring X-ray prohibited item detection from a close-set to an open-set paradigm by implementing open-vocabulary object detection, which enables the detection of objects from unseen novel categories using a detector trained on labeled base categories. Importantly, our method achieves this without requiring expensive annotations or costly training, making it well-suited for real-world X-ray security inspection scenarios. To address the challenge of sharp performance drops caused by domain shift when directly applying CLIP to distillation-based OVOD methods in X-ray domain, we propose X-ray feature adapter, which involves three submodules to adapt CLIP to the X-ray OVOD task. Specifically, adapter submodules effectively bridge domain gap by integrating newly learned specific knowledge from the X-ray domain with original knowledge in CLIP. Extensive experiments conducted on PIXray and PIDray datasets demonstrate that forming OVXD by applying X-ray feature adapter to CLIP within OVOD framework can significantly surpasses baseline OVOD methods. Additionally, we demonstrate generalization performance of proposed OVXD by directly transferring detector trained on SIXray to PIXray and PIDray datasets. As part of our future research, we plan to expand existing dataset and release the first prohibited item dataset with image-caption pairs to further explore latent visual-language concepts for detection in X-ray security inspection scenarios. 

\bibliographystyle{IEEEtran}
\bibliography{IEEEabrv, reference}
\vfill

\end{document}